\documentclass[conference,9pt]{IEEEtran}
\IEEEoverridecommandlockouts

\usepackage{cite}
\usepackage{amsmath,amssymb,amsfonts}
\usepackage{graphicx}
\usepackage{placeins}
\usepackage{textcomp}
\usepackage{xcolor}
\usepackage{url}

\def\BibTeX{{\rm B\kern-.05em{\sc i\kern-.025em b}\kern-.08em
    T\kern-.1667em\lower.7ex\hbox{E}\kern-.125emX}}

\begin{document}
\title{REDDIT: Forgetting-Resistant Correction of Timestamp Drift in ASR via Replay-Based Distribution Editing}

\author{
    \IEEEauthorblockN{
        Cheng-Kang Chou${}^{*1}$, 
        Ming-Douo Tchouang${}^{*1}$, 
        Ke-Han Lu${}^{1}$, 
        Chan-Jan Hsu${}^{2}$, 
        Hung-yi Lee${}^{3}$
    }
    \IEEEauthorblockA{\small
        ${}^{1}$National Taiwan University \;\;
        ${}^{2}$Carnegie Mellon University \;\;
        ${}^{3}$NTU Artificial Intelligence Center of Research Excellence (NTU AI-CoRE)\\[0pt]
        \scriptsize\texttt{r14942095@ntu.edu.tw, b08901172@g.ntu.edu.tw, d12942024@ntu.edu.tw, a24998667@gmail.com, tlkagkb93901106@gmail.com}\vspace{-6pt}
    }
    \thanks{* Equal contribution.}
}

\maketitle
\begin{abstract}
Autoregressive ASR systems can emit timestamps as decoded tokens, but these timestamps may drift across long non-speech spans even when transcripts remain plausible. We study this failure across 15 timestamp-producing ASR and audio-language systems using controlled benchmarks with inserted non-speech gaps. We propose \emph{REDDIT} (REplay-based Distribution eDITing), a two-stage, forgetting-resistant post-training framework. Stage 1 edits timestamp targets under replayed decoder contexts while anchoring the student's output distributions at non-timestamp positions to those of a frozen teacher; Stage 2 applies a short refinement conditioned on decoder prefixes containing corrected timestamp tokens. Using 34.9 hours of correction audio and updating only 1.6\% of Whisper-tiny parameters, REDDIT raises mean temporal overlap on synthetic clips containing a long non-speech interval from 38.7\% to 95.0\%. In out-of-domain evaluations, REDDIT reduces mean timestamp deviation on gap-augmented English and Mandarin speech from 2752 to 223 ms; on unmodified English speech, its MER is 41.3\%, compared with 37.0\% for the base model and 524.2\% for decoder SFT. Evaluations on real-world broadcast audio further show that REDDIT transfers to naturally occurring non-speech intervals. On clips from \emph{The Daily Show}, REDDIT reduces  AAS from 3010 to 1252 ms and MER from 30.8\% to 24.4\%, with corresponding reductions on Mandarin broadcast audio.

\end{abstract}

\section{Introduction}
\label{sec:introduction}

Automatic speech recognition (ASR) and audio-language models increasingly need to answer not only \emph{what} was said, but also \emph{when} it was said. Autoregressive speech systems expose this interface by generating time as part of the decoded output, either as timestamp tokens in ASR models such as Whisper or as word-level and text-form temporal predictions in speech-aware language models \cite{whisper,word_level_timestamp_generation,insync_asr,timeaudio}. This self-generated timestamp interface avoids the need for a separate VAD, frame-level aligner, or forced-alignment module at inference time.

This convenience introduces a failure mode that transcript-centric metrics can miss. Speech evaluation already extends beyond lexical content, e.g., tonality \cite{toxictone}; here, the missing dimension is temporal placement. Because the time axis is generated by the same decoder that emits text, a model can produce plausible content while assigning it to the wrong region of the audio. We focus on \emph{non-speech-induced timestamp drift}: when speech follows a long silent or non-speech span, self-generated timestamps may place the following speech too early, too late, or under a globally displaced time axis, rendering subtitle timing unusable despite having an acceptable word error rate or task-level transcript quality.

One possible cause is a mismatch between common training data and long-gap deployment conditions: timestamped ASR examples often begin near speech onset, and VAD-based trimming
  further increases speech density, potentially limiting exposure to long non-speech prefixes and strengthening an early-timestamp prior \cite{whisperx}.

Post-hoc VAD, forced-alignment, and attention-based pipelines can refine boundaries, but they add inference-time components and do not correct the model's native timestamp generation \cite{whisperx}. Timestamp drift also differs from lexical hallucination: hallucination produces unsupported words, whereas drift places supported words at the wrong time.

Naïve timestamp-corrected fine-tuning can improve target alignment while degrading recognition on non-target ASR data. The task is therefore not merely timestamp supervision, but forgetting-resistant temporal editing: moving timestamps to the correct positions while preserving the decoder's original non-timestamp distribution.

We first benchmark non-speech-induced drift across 15 evaluated timestamp-producing ASR and audio-language systems using controlled gap and long-gap evaluations. For the intervention study, we focus on Whisper-tiny and evaluate transfer to unspliced English and Mandarin broadcast audio. This setting focuses on Whisper-style timestamp-token ASR, where timestamp positions are explicit and token-level distribution editing is well-defined.

We propose \emph{REDDIT} (REplay-based Distribution eDITing), a two-stage post-training framework for models that transcribe adequately but exhibit non-speech-induced timestamp drift. Stage 1 edits timestamp targets under cached decoder replay contexts while anchoring the student's non-timestamp output distributions to those of a frozen teacher; Stage 2 applies a short refinement conditioned on decoder prefixes containing corrected timestamp tokens. In our Whisper-tiny experiments, \emph{REDDIT} uses 34.9 hours of automatically constructed correction audio, updates only 0.59M (1.6\%) parameters, and requires no inference-time alignment or post-processing.

Our contributions are:
\begin{itemize}
    \item We identify non-speech-induced timestamp drift as a distinct failure mode of self-generated timestamps in ASR and audio-language systems, where plausible transcripts can be temporally misplaced across non-speech spans.
    \item We establish a diagnostic evaluation framework to benchmark timestamp-producing ASR systems on speech audio interleaved with challenging non-speech gaps, isolating joint alignment and recognition failures.
    \item We show that naïve timestamp-corrected fine-tuning can cause severe forgetting on non-target ASR data, motivating timestamp adaptation as a model-editing problem.
    \item We propose \emph{REDDIT}, a two-stage pipeline that constructs correction examples without human transcripts or human timestamp annotations, edits timestamp behavior under cached replay context, and refines the corrected behavior while preserving non-target recognition.
\end{itemize}

\begin{figure*}[t]
  \centering
  \includegraphics[width=0.9\textwidth]{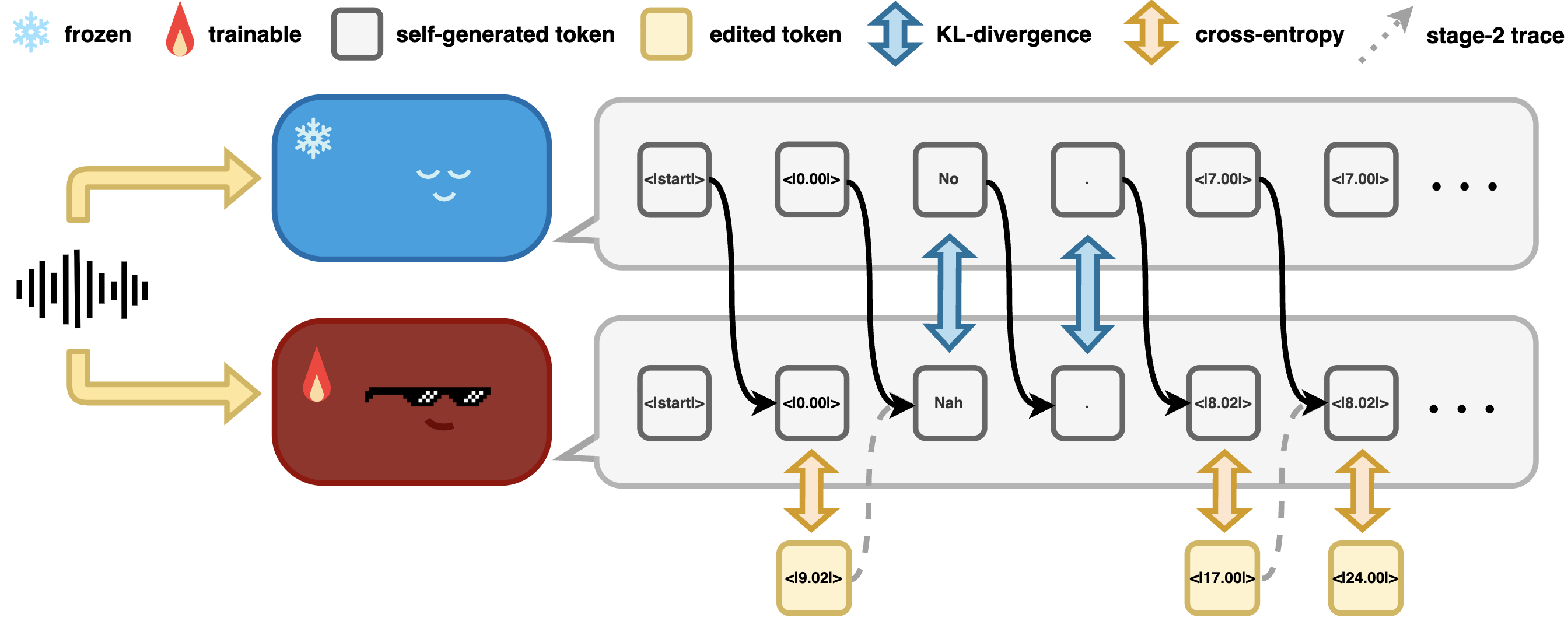}
  \caption{Overview of the REDDIT framework. A cached base-model replay sequence serves as the decoder context. Timestamp targets are edited based on known synthetic offsets and optimized via cross-entropy loss, while non-timestamp positions are trained to match the corresponding frozen teacher distribution via KL divergence. The Stage-2 trace uses the edited tokens as the teacher-forcing prefix to refine the Stage-1 checkpoint.}
  \label{fig:reddit_overview}
\end{figure*}

\section{Related Work}
\label{sec:related}

\subsection{Generated Timestamp Prediction}

Autoregressive ASR systems increasingly expose time as part of the decoded sequence. Whisper emits timestamp tokens together with text, enabling timestamped transcription without a separate frame-level alignment model \cite{whisper}; later work extends this interface to explicit word-level timestamp generation \cite{word_level_timestamp_generation}. These systems make time a native model output rather than metadata recovered after decoding.

The closest concurrent work is In-Sync, which adapts a speech-aware language model for joint ASR and word-level timestamp prediction \cite{insync_asr}. It uses speech-length augmentation, timestamp embedding regularization, and reduced teacher forcing to address heavy-tailed timestamp token values, missing monotonic structure, and error propagation. In-Sync predicts end-of-word timestamps and reports that an explicit silence token degraded performance, leaving silence modeling for future work. Our diagnostic setting instead follows the start/end segment timecode format commonly used by timestamp-token ASR and LALM-style temporal outputs: non-speech is represented by the gap between adjacent end and start timecodes, as well as pre- and post-speech offsets. REDDIT targets this non-speech dimension in a post-training model-editing setting, where an already capable timestamp-producing ASR model is edited and forgetting becomes part of the objective.

Broader speech and audio-language benchmarks have also moved beyond conventional ASR toward instruction-following, regression, sequence generation, and general audio understanding tasks \cite{superb,ml_superb,dynamic_superb_phase2,hear,air_bench,audiobench,mmau,audio_marathon,not_in_sync,spotsound,chronosaudio,star_bench,audioset,audiocaps,clotho,clotho_aqa}. Our benchmark is complementary: rather than measuring overall spoken-language capability, it isolates temporal reliability in systems that emit timestamps or timestamp-like temporal outputs.

\subsection{Post-Hoc and Internal Alignment}

Many timestamping systems align transcript units after, or alongside, decoding. Forced aligners such as the Montreal Forced Aligner use acoustic and pronunciation constraints \cite{mfa,kaldi,forced_alignment_comparison,ctc_segmentation,llm_forced_aligner}, and WhisperX combines VAD segmentation with forced phoneme alignment for long-form Whisper transcription \cite{whisperx}. CrisperWhisper fine-tunes Whisper for more verbatim transcription, but its word-level timestamps still rely on dynamic time warping over decoder cross-attention, attention supervision, and pause heuristics \cite{crisper_whisper}. Useful alignments can also be extracted from selected Whisper cross-attention heads without additional training \cite{whisper_internal_aligner}.

These methods ask where transcript units should be placed on the timeline. REDDIT asks whether the decoder's own generated timestamp-token distribution can be edited so that corrected timestamps are produced natively at inference time, without DTW, forced alignment, VAD segmentation, or attention-head selection.

\subsection{Non-Speech Robustness}

Non-speech audio can trigger unsupported text, repeated phrases, or boilerplate outputs in autoregressive ASR, especially when language-model priors dominate weak acoustic evidence. Prior work studies Whisper hallucinations under non-speech conditions and small targeted interventions for hallucination reduction \cite{whisper_nonspeech_hallucination,calm_whisper,lost_in_transcription,aha_bench,whisper_hallu_detection,whisper_hallu_steering}. Timestamp drift is related but not identical: the words may be acoustically supported, yet assigned to the wrong interval.

\subsection{Forgetting-Resistant Editing}

Adapting a pretrained speech model can improve a target behavior while damaging behavior outside the adaptation set. Speech adaptation work often uses self-refining synthetic data, domain-adaptive simulation, or parameter-efficient updates to improve robustness under limited data or domain mismatch \cite{self_refining_tts_asr,channel_adaptive_gan,unseen_language_adaptation,adapter_tuning,lora,prefix_tuning,prompt_tuning,bitfit,ia3}. Distillation and learning-without-forgetting objectives preserve prior behavior by matching teacher distributions or retaining previous-task responses \cite{distillation,lwf,ewc,gem,dark_experience_replay}; speech-language evaluation further shows that speech-centric adaptation can harm broader instruction-following ability \cite{speech_ifeval}. REDDIT applies this idea at the token level: timestamp positions receive edited temporal supervision, while non-timestamp positions replay the frozen base distribution under the same cached replay context. This separates temporal correction from recognizer preservation, making the method closer to distribution editing than ordinary timestamp fine-tuning.

\section{Method}
\label{sec:method}

\subsection{Problem Formulation}

Let $x$ be an input speech example and $F_{\theta_0}$ be a frozen pretrained
autoregressive ASR model. In timestamp mode it generates a sequence over the
vocabulary $\mathcal{V} = \mathcal{V}_{\mathrm{text}} \sqcup \mathcal{V}_{\mathrm{time}}$,
the disjoint union of lexical and timestamp tokens, where consecutive timestamp
tokens delimit decoded segment boundaries.

We define \emph{model-generated timestamp drift} as a mismatch between the decoded
and acoustic timelines: around non-speech gaps, the decoder can place acoustically
supported words in the wrong time region, a failure distinct from lexical
hallucination.

REDDIT corrects this drift from a human-annotation-free correction set of $N$ samples,
\begin{equation}
    \mathcal{D}_{\mathrm{corr}} = \{(x_i, b_i)\}_{i=1}^{N},
\end{equation}
where $b_i = (b_{i,1}, \dots, b_{i,2K_i}) \in \mathcal{V}_{\mathrm{time}}^{2K_i}$ is a
sequence of timestamp tokens encoding the corrected start/end boundaries of the
$K_i$ speech spans in $x_i$ (Sec.~\ref{subsec:data}). No human transcript or
timestamp annotation is required: pseudo-text and output distributions are supplied
by $F_{\theta_0}$ itself, while $b_i$ supplies only the timestamp edit targets.
Figure~\ref{fig:reddit_overview} summarizes how these targets are used across the
two stages.

\subsection{Stage 1: Cached Replay Context and Target Editing}

For each example we decode once with the frozen base model and cache the trajectory,
\begin{equation}
    \tilde{y}_i = F_{\theta_0}(x_i) = (\tilde{y}_{i,1}, \dots, \tilde{y}_{i,T_i}),
\end{equation}
keeping it fixed throughout training. Let
\begin{equation}
    \mathcal{T}_i = \{\, t \in \{1,\dots,T_i\} : \tilde{y}_{i,t} \in \mathcal{V}_{\mathrm{time}} \,\}
\end{equation}
be the set of timestamp positions in the replay, with elements ordered as
$t_1 < \dots < t_{|\mathcal{T}_i|}$. Filtering (Sec.~\ref{subsec:data}) guarantees
$|\mathcal{T}_i| = 2K_i$, so the $j$-th replay timestamp aligns with the $j$-th
boundary token $b_{i,j}$. We edit the time axis by overwriting each replay timestamp
with its corrected target, leaving every other position untouched:
\begin{equation}
    y^{\mathrm{tar}}_{i,t} =
    \begin{cases}
        b_{i,j}, & t = t_j,\ j = 1,\dots,2K_i, \\[2pt]
        \tilde{y}_{i,t}, & t \notin \mathcal{T}_i.
    \end{cases}
\end{equation}

In Stage 1, the edited timestamps act \emph{only as targets}: they are never fed back
as the decoder input for the next step. The decoder is always conditioned on the
cached replay prefix $\tilde{y}_{i,<t}$, so standard teacher forcing is repurposed
purely as a context-injection mechanism. Let $F_\theta$ denote the student, a
trainable copy of the base model initialized from $\theta_0$, with output
distribution $p_\theta$. Timestamp supervision then optimizes the student under the
replay prefix, $p_\theta(y^{\mathrm{tar}}_{i,t}\mid \tilde{y}_{i,<t}, x_i)$, rather
than the edited-prefix objective
$p_\theta(y^{\mathrm{tar}}_{i,t}\mid y^{\mathrm{tar}}_{i,<t}, x_i)$. This is an offline
approximation to extreme scheduled sampling \cite{scheduled_sampling}: the student is
not rolled out live.

\subsection{Stage 1 Replay Editing Objective}

Student and teacher are scored under the same replay context
\begin{equation}
    c_{i,t} = (\tilde{y}_{i,<t}, x_i).
\end{equation}
Let $\mathcal{T}_i$ be the timestamp positions defined above and $\mathcal{X}_i$ the
remaining valid (non-special, non-padding) positions. REDDIT edits the time axis on
$\mathcal{T}_i$ with cross-entropy,
\begin{equation}
    \mathcal{L}_{\mathrm{time}}
    = -\frac{1}{\sum_i |\mathcal{T}_i|}
      \sum_{i}\sum_{t\in\mathcal{T}_i}
      \log p_\theta\!\left(y^{\mathrm{tar}}_{i,t}\mid c_{i,t}\right),
\end{equation}
and preserves base behavior on $\mathcal{X}_i$ by matching the frozen teacher under
the identical context,
\begin{equation}
    \mathcal{L}_{\mathrm{text}}
    = \frac{1}{\sum_i |\mathcal{X}_i|}
      \sum_{i}\sum_{t\in\mathcal{X}_i}
      D_{\mathrm{KL}}\!\left(p_{\theta_0}(\cdot\mid c_{i,t}) \,\|\, p_\theta(\cdot\mid c_{i,t})\right),
\end{equation}
where $p_{\theta_0}(\cdot\mid c_{i,t})$ is obtained from the frozen base model under
the replay context. The Stage-1 objective is
\begin{equation}
    \mathcal{L}_{\mathrm{S1}}
    = \lambda_{\mathrm{time}}\mathcal{L}_{\mathrm{time}}
    + \lambda_{\mathrm{text}}\mathcal{L}_{\mathrm{text}}.
\end{equation}
Here, $\lambda_{\mathrm{time}}$ and $\lambda_{\mathrm{text}}$ are hyperparameters balancing the two loss terms.

Stage 1 edits timestamp targets without rewriting the temporal context. Because the decoder prefix is always the cached, potentially drifted replay, the learned timestamp transitions remain conditioned on an offline proxy for model-generated inference prefixes.

\subsection{Stage 2: Edited-Prefix Refinement}

The full REDDIT pipeline then runs a short Stage-2 refinement from the Stage-1 checkpoint. Let $\theta_1$ denote the Stage-1 parameters. We form an edited-prefix context from $y^{\mathrm{tar}}_i$,
\begin{equation}
    c^{\mathrm{edit}}_{i,t} = (y^{\mathrm{tar}}_{i,<t}, x_i).
\end{equation}
Stage 2 keeps the same correction examples and timestamp targets, and uses the same timestamp-CE and non-timestamp-KL loss forms as Stage 1. The difference is contextual: both terms are computed under $c^{\mathrm{edit}}_{i,t}$, and the non-timestamp KL teacher is the frozen Stage-1 checkpoint $F_{\theta_1}$ rather than the frozen base model $F_{\theta_0}$. This stage adds no annotation or inference-time modules; it consolidates the corrected timestamp transitions after Stage 1 has exposed the model to drifted replay prefixes.

\subsection{Model}

In both stages, we freeze the encoder, token embeddings, decoder self-attention, decoder
feed-forward layers, and output projection, and update only last cross-attn + LNs.
Here, LNs denote the layer norms in the last decoder block and the final decoder
layer norm.

\subsection{Data Preparation}
\label{subsec:data}

We describe construction for a generic example and drop the index $i$. Correction
examples are built by VAD-trimming clean speech spans and inserting sampled
non-speech audio before, between, and after them \cite{rvad,whisperx}, yielding
\emph{pre-speech}, \emph{inter-speech}, and \emph{post-speech} gaps. Let $a_k$ denote
a VAD-trimmed speech span and $g_k$ a non-speech segment; a synthetic example with
$K$ spans is
\begin{equation}
    x = g_0 \oplus a_1 \oplus g_1 \oplus \cdots \oplus a_K \oplus g_K,
\end{equation}
with $\oplus$ audio concatenation. The start/end times of the $k$-th span follow
directly from the inserted durations,
\begin{equation}
    \tau_k^{s} = \sum_{r=0}^{k-1}|g_r| + \sum_{r=1}^{k-1}|a_r|,
    \qquad
    \tau_k^{e} = \tau_k^{s} + |a_k|,
\end{equation}
and, since the spans are placed by construction, are exact and need no annotation.
Quantizing each boundary to its nearest timestamp token via $q(\cdot)$ yields the
edit target
\begin{equation}
    b = \big(q(\tau_1^{s}), q(\tau_1^{e}), \dots, q(\tau_K^{s}), q(\tau_K^{e})\big)
      \in \mathcal{V}_{\mathrm{time}}^{2K}.
\end{equation}
The frozen base model supplies pseudo-text and cached replay timestamps for the same
$x$. Before timestamp editing, we pre-filter the cached teacher replays offline to
remove hallucinations, repetitions, boilerplate, empty or unusable text, and
structurally inconsistent timestamps. Each retained replay therefore has exactly
$2K$ well-ordered timestamp tokens to be overwritten by $b$.

\providecommand{\best}[1]{#1}
\providecommand{\worst}[1]{#1}
\renewcommand{\best}[1]{\textbf{\textcolor{green!50!black}{#1}}}
\renewcommand{\worst}[1]{\textbf{\textcolor{red}{#1}}}
\providecommand{\rawrate}[2]{#2}
\providecommand{\pairrawrate}[4]{\rawrate{#1}{#2} / \rawrate{#3}{#4}}
\section{Experiments}
\label{sec:experiments}

We evaluate three questions: how severe generated timestamp drift is under controlled non-speech gaps, whether REDDIT can correct it with limited correction data, and whether the correction preserves non-target ASR behavior.

\subsection{Evaluation Data}

We use Common Voice zh-TW audio \cite{common_voice} for targeted correction data and in-domain evaluation. The \emph{Gap} split (\texttt{test\_gap}) concatenates two or three utterances with inserted non-speech regions, testing multi-segment temporal tracking. The \emph{Long-Gap} split (\texttt{test\_long\_gap}) contains one utterance with a long leading or trailing non-speech region, isolating onset and offset anchoring.

Both in-domain evaluation splits contain 5{,}105 examples capped at 30~s. Reference timestamps are exact because they are computed from the splicing schedule. Non-speech audio is sampled from a held-out non-speech pool, and test non-speech and speech sources are disjoint from training. Table~\ref{tab:target-data-summary} summarizes the targeted correction and in-domain benchmark splits.

\begin{table}[!t]
  \centering
  \scriptsize
  \caption{Target correction and in-domain timestamp benchmark data from \texttt{WTFO/cmv\_time\_shift}. Hours are computed from audio and timestamp labels.}
  \label{tab:target-data-summary}
  \resizebox{\columnwidth}{!}{
  \begin{tabular}{lccccc}
  \hline
  Role & Samples & Audio (h) & Speech (h) & Non-speech (h) & Speech spans \\
  \hline
  Train & 7{,}367 & 34.9 & 7.3 & 27.6 & 1--3 \\
  Gap eval & 5{,}105 & 21.3 & 7.0 & 14.3 & 2--3 \\
  Long-gap eval & 5{,}105 & 30.5 & 3.1 & 27.5 & 1 \\
  \hline
  \end{tabular}
  }
\end{table}

To test retention and transfer, we also evaluate ASCEND-ZH, ASCEND-EN, ASCEND-MIXED \cite{ascend}, and CommonVoice-EN-min \cite{common_voice} in two versions: the original no-gap split and a gap stress split constructed with the same gap-insertion procedure. Table~\ref{tab:ood-data-summary} summarizes these OOD evaluation sets.

\begin{table}[!t]
  \centering
  \scriptsize
  \caption{Out-of-domain retention and mixed-gap evaluation data. N is sample count; h is duration in hours. Gap sets are gap-stressed counterparts of the no-gap sets. CV-en-min denotes CommonVoice-EN-min.}
  \label{tab:ood-data-summary}
  \resizebox{\columnwidth}{!}{
  \begin{tabular}{lccccc}
  \hline
  Dataset & No-gap N & No-gap h & Gap N & Gap h & Non-speech h \\
  \hline
  ASCEND-zh & 578 & 0.3 & 578 & 2.6 & 2.1 \\
  ASCEND-en & 214 & 0.1 & 214 & 1.0 & 0.8 \\
  ASCEND-mix & 373 & 0.4 & 373 & 2.0 & 1.3 \\
  CV-en-min & 2{,}997 & 3.2 & 2{,}997 & 16.7 & 10.8 \\
  \hline
  Total & 4{,}162 & 4.0 & 4{,}162 & 22.4 & 15.0 \\
  \hline
  \end{tabular}
  }
\end{table}
 To evaluate transfer beyond synthetically constructed gaps, we build two natural-audio evaluation sets from unspliced broadcast excerpts: \emph{The Daily Show} in English
  (1,090 clips) and \emph{The Night Night Show} in Mandarin (732 clips). Speech boundaries are obtained using VAD, while reference text is taken from uploader-provided YouTube captions; YouTube auto-generated caption tracks are excluded. Each set contains multi-span clips with naturally occurring inter-speech intervals and single-span clips
  with long leading or trailing non-speech. The intervals include silence, room tone, laughter, applause, and music. All clips are at most 30~s and are decoded within a single
  window.

\subsection{Systems and Baselines}

The diagnostic benchmark covers timestamp-producing ASR and audio-language systems, including Whisper-family models \cite{whisper}, Distil-Whisper \cite{distil_whisper}, Qwen audio/omni models \cite{qwen2_audio,qwen25_omni,qwen3_omni}, MOSS-Audio models \cite{moss_audio}, VibeVoice-ASR \cite{vibevoice_asr}, and DeSTA-style audio-language models \cite{desta25_audio}. Intervention experiments focus on Whisper-tiny to isolate whether a small timestamp-token ASR update can repair drift without adding inference-time alignment.

Baselines include base Whisper-tiny, supervised timestamp-corrected fine-tuning (SFT), timestamp-only fine-tuning, edited-context replay, REDDIT Stage 1, REDDIT Full, and an In-Sync-inspired reduced teacher forcing (RTF) variant adapted to Whisper-tiny. RTF corrupts the previous timestamp input with probability $p_{\mathrm{rtf}}=0.2$ at timestamp-target positions while leaving text inputs unchanged. This is a baseline inspired by In-Sync \cite{insync_asr}, not a full reimplementation of its length augmentation, embedding regularization, or end-of-word timestamp formulation. REDDIT Stage 1 isolates replay-conditioned distribution editing; REDDIT Full is the complete two-stage pipeline, which starts from the Stage-1 checkpoint and further trains on the same correction data with edited-prefix refinement.

Table~\ref{tab:method-config} summarizes the trainable scope used by the Whisper-tiny interventions.

\begin{table}[!t]
  \centering
  \scriptsize
  \caption{Configuration of compared Whisper-tiny methods. LNs denote the last decoder-block layer norms plus the final decoder layer norm; REDDIT Full uses the same scope in both stages.}
  \label{tab:method-config}
  \resizebox{\columnwidth}{!}{
  \begin{tabular}{lcc}
  \hline
  Method & Trainable scope & Trainable params \\
  \hline
  Base Whisper-tiny & 0 & 0 \\
  SFT & last cross-attn + LNs & 0.59M (1.6\%) \\
  SFT timestamp-only & last cross-attn + LNs & 0.59M (1.6\%) \\
  SFT decoder & decoder & 29.38M (77.8\%) \\
  Edited-time replay & last cross-attn + LNs & 0.59M (1.6\%) \\
  Edited-time replay decoder & decoder & 29.38M (77.8\%) \\
  Reduced Teacher Forcing & last cross-attn + LNs & 0.59M (1.6\%) \\
  Replayed Reduced Teacher Forcing & last cross-attn + LNs & 0.59M (1.6\%) \\
  \textbf{REDDIT Stage 1 (Ours)} & last cross-attn + LNs & 0.59M (1.6\%) \\
  \textbf{REDDIT Full (Ours)} & last cross-attn + LNs & 0.59M (1.6\%) \\
  \hline
  \end{tabular}
  }
\end{table}
 
\subsection{Post-Training Setup}

All timestamp intervention runs use 34.9 hours of targeted correction audio after cached-replay pre-filtering. Stage 1 uses the same training budget as the single-stage baselines, and REDDIT Full adds a short Stage-2 refinement on the same correction data. We select the reported full-pipeline checkpoint from this short refinement trajectory by validation behavior rather than treating a fixed step count as part of the method. We freeze the encoder, token embeddings, decoder self-attention, decoder feed-forward layers, and output projection, updating only last cross-attn + LNs. This scope contains 0.59M of 37.76M Whisper-tiny parameters (1.6\%). The reported Whisper-large-v3 REDDIT runs use the same last-layer scope, corresponding to 6.57M of 1.54B parameters (0.43\%).

We use AdamW, learning rate $1\times10^{-5}$ after 10 warmup steps, and batch size 64 for all post-training runs. Single-stage baselines and REDDIT Stage 1 use the same duration. REDDIT Stage 1 uses $\lambda_{\mathrm{time}}=1$ and $\lambda_{\mathrm{text}}=5$. Stage 2 uses the same timestamp-CE and non-timestamp-KL loss forms under edited-prefix context, with the frozen Stage-1 checkpoint as the KL teacher. The replay KL uses temperature 1.0, no top-$k$ truncation, and online teacher logits under the corresponding replay or edited-prefix context.

\subsection{Evaluation Metrics}
\label{sec:metrics}

We report a compact diagnostic set in the main paper tables.

\paragraph{Temporal overlap}
For a matched hypothesis segment $p=[s_p,e_p]$ and reference segment $r=[s_r,e_r]$,
\[
\mathrm{tIoU}(p,r)=
\frac{|p \cap r|}{|p \cup r|}.
\]
We report mean tIoU (mIoU) over matched pairs and omit match rate from the main tables because textual matching alone does not imply temporal overlap.

\paragraph{Boundary error.}
\emph{Start MAE} and \emph{End MAE} measure onset and offset errors over matched segments, in seconds. Start MAE captures subtitles appearing too early or late after non-speech gaps; end MAE captures subtitles ending too early or extending into silence.

\paragraph{Token-level alignment.}
We report AAS in milliseconds:
\[
\mathrm{AAS}=\frac{\mathrm{ASTD}+\mathrm{AETD}}{2},
\]
where ASTD and AETD are start- and end-time deviations over edit-distance paired text units. Insertions and deletions are excluded. MAL is the percentage of samples without a valid paired timed token, preventing low AAS from hiding missing alignments.

\paragraph{Large drift and hallucination.}
Metrics \emph{Drift$>$5s} and \emph{Drift$>$10s} are the percentages of matched segments with absolute temporal displacement above 5 or 10 seconds. Hallucination Rate is a file-level diagnostic based on common hallucination phrases and repeated $n$-grams for $n\in\{3,4,5\}$, following prior non-speech hallucination analyses \cite{whisper_nonspeech_hallucination,calm_whisper}; it is reported separately because supported speech can be temporally misplaced without being hallucinated.

\paragraph{Retention.}
For non-target ASR retention, we report MER according to benchmark convention. These text metrics are not timestamp metrics; they measure whether timestamp adaptation preserves recognition behavior outside the correction set.

\subsection{Ablations}

Ablations isolate timestamp supervision, replay, decoder context, and reduced teacher forcing. The key comparison is whether corrected timestamps are learned under teacher-forced edited prefixes or under cached base-model replay prefixes, with non-timestamp behavior preserved by distribution replay.
 \begin{table}[!t]
  \centering
  \scriptsize
  \caption{Results on natural broadcast audio. Each paired entry reports \emph{The Daily Show} (en;
  $N=1{,}090$) / \emph{The Night Night Show} (zh; $N=732$).}
 \label{tab:natural-broadcast}
  \resizebox{\columnwidth}{!}{
  \begin{tabular}{lccc}
  \hline
  Method
  & AAS (ms) $\downarrow$
  & Drift$>$5s (\%) $\downarrow$
  & MER (\%) $\downarrow$ \\
  \hline
  Base Whisper-tiny
  & 3010 / 3870
  & 44.3 / 40.9
  & 30.8 / 71.5 \\
  SFT decoder
  & 1259 / \best{2008}
  & 9.8 / 15.0
  & 92.2 / 61.9 \\
  \textbf{REDDIT Full (Ours)}
  & \best{1252} / 2257
  & \best{8.6} / \best{14.9}
  & \best{24.4} / \best{58.2} \\
  \hline
  \end{tabular}
  }
  \end{table}
% \begin{table}[!t]
%   \centering
%   \scriptsize
%   \caption{Results on natural broadcast audio. Each paired entry reports \emph{The Daily Show} (en;
%   $N=1{,}090$) / \emph{The Night Night Show} (zh; $N=732$).}
%  \label{tab:natural-broadcast}
%   \resizebox{\columnwidth}{!}{
%   \begin{tabular}{lccc}
%   \hline
%   Method
%   & AAS (ms) $\downarrow$
%   & Drift$>$5s (\%) $\downarrow$
%   & MER (\%) $\downarrow$ \\
%   \hline
%   Base Whisper-tiny
%   & 3010 / 3870
%   & 44.3 / 40.9
%   & 30.8 / 71.5 \\
%   SFT decoder
%   & 1259 / \best{2008}
%   & 9.8 / 15.0
%   & 92.2 / 61.9 \\
%   \textbf{REDDIT Full (Ours)}
%   & \best{1252} / 2257
%   & \best{8.6} / \best{14.9}
%   & \best{24.4} / \best{58.2} \\
%   \hline
%   \end{tabular}
%   }
%   \end{table}
\section{Results}
\label{sec:results}

\providecommand{\best}[1]{#1}
\providecommand{\worst}[1]{#1}
\renewcommand{\best}[1]{\textbf{\textcolor{green!50!black}{#1}}}
\renewcommand{\worst}[1]{\textbf{\textcolor{red}{#1}}}
\providecommand{\rawrate}[2]{#2}
\providecommand{\pairrawrate}[4]{\rawrate{#1}{#2} / \rawrate{#3}{#4}}

% \subsection{Non-Speech Gaps Expose Timestamp Drift}

% Table~\ref{tab:diagnostic-main} shows that transcript quality does not guarantee temporal grounding. Whisper-family models often produce recognizable text while long-gap start MAE and large-drift rates increase sharply; for example, Whisper-tiny mIoU drops from 63.0\% to 38.7\%, while start MAE rises from 1.30~s to 7.34~s. The strongest lexical systems are not automatically the strongest temporal systems: Whisper-large-v3 has the best gap MER among Whisper models, but it also has the worst gap Drift$>$5s rate. These results support evaluating timestamp drift as an acoustic-grounding error rather than as a byproduct of transcript quality. The table also separates drift from hallucination: large temporal displacements measure misplaced supported speech, whereas hallucination rate measures unsupported loops or boilerplate.

\begin{table*}[!t]
  \centering
  \scriptsize
  \caption{Timestamp and non-speech results for all retained model families. Each paired cell reports
  gap / long-gap. Drift and hallucination columns are file-level rates, with raw counts retained in the source.
  Green and red mark best and worst values.}
  \label{tab:diagnostic-main}
  \setlength{\tabcolsep}{2.4pt}
  \resizebox{0.98\textwidth}{!}{
  \begin{tabular}{@{}lccccccccc@{}}
  \hline
  System & mIoU \% $\uparrow$ & MER \% $\downarrow$ & AAS (ms) $\downarrow$ & MAL \% $\downarrow$ & Start
  MAE (s) $\downarrow$ & End MAE (s) $\downarrow$ & Drift$>$5s \% $\downarrow$ & Drift$>$10s \%
  $\downarrow$ & Halluc. \% $\downarrow$ \\
  \hline
  MOSS-Audio-4B & \best{81.0} / \best{85.0} & 39.5 / 34.6 & 521 / 377 & \best{0.0} /
  \best{0.1} & \best{0.23} / \best{0.12} & 0.38 / 0.28 & \pairrawrate{122/5105}{2.4}{15/5105}{0.3}
  & \pairrawrate{4/5105}{0.1}{5/5105}{0.1} & \pairrawrate{299/5105}{\best{5.9}}{25/5105}
  {\best{0.5}} \\
  MOSS-Audio-8B & 76.3 / 63.5 & 31.7 / 32.5 & 468 / 2849 & 0.5 / 1.7 & 0.63 / 3.59 & 0.36 / 0.29 &
  \pairrawrate{572/5105}{11.2}{1304/5105}{25.5} & \pairrawrate{5/5105}{0.1}{807/5105}{15.8} &
  \pairrawrate{357/5105}{7.0}{158/5105}{3.1} \\
  Qwen3-Omni-30B & 76.9 / 81.4 & 22.0 / \best{9.3} & 814 / \best{300} & 3.1 / 1.3 & 0.31 /
  0.26 & \best{0.31} / 0.23 & \pairrawrate{35/5105}{0.7}{13/5105}{0.3} & \pairrawrate{1/5105}
  {0.0}{8/5105}{0.2} & \pairrawrate{487/5105}{9.5}{123/5105}{2.4} \\
  VibeVoice-ASR & 71.2 / 84.1 & 42.5 / 48.1 & \best{465} / 568 & 3.8 / 12.8 & 0.87 / 0.48 & 0.80 /
  0.92 & \pairrawrate{485/5105}{9.5}{46/5105}{0.9} & \pairrawrate{41/5105}{0.8}{20/5105}{0.4}
  & \pairrawrate{587/5105}{11.5}{735/5105}{14.4} \\
  Whisper-tiny & 63.0 / 38.7 & 79.9 / \worst{347.2} & 951 / 4806 & 5.4 / 17.5 & 1.30 / 7.34 & 0.46 /
  0.28 & \pairrawrate{777/5105}{15.2}{1883/5105}{36.9} & \pairrawrate{19/5105}{0.4}{1365/5105}
  {26.7} & \pairrawrate{664/5105}{13.0}{982/5105}{19.2} \\
  Whisper-base & 61.8 / 39.8 & 56.9 / 232.8 & 1063 / 5171 & 3.0 / 13.2 & 1.48 / 7.50 & 0.42 /
  \best{0.21} & \pairrawrate{1117/5105}{21.9}{1876/5105}{36.7} & \pairrawrate{24/5105}{0.5}
  {1298/5105}{25.4} & \pairrawrate{523/5105}{10.2}{906/5105}{17.7} \\
  Whisper-medium & 64.8 / 45.2 & 97.5 / 324.5 & 1209 / 4544 & 2.7 / 5.9 & 1.16 / 6.00 & 0.39 / 0.47
  & \pairrawrate{681/5105}{13.3}{1552/5105}{30.4} & \pairrawrate{46/5105}{0.9}{1189/5105}{23.3} &
  \pairrawrate{712/5105}{13.9}{1454/5105}{28.5} \\
  Whisper-large-v2 & 65.2 / 45.8 & 64.9 / 139.6 & 1334 / 4687 & 1.5 / 3.6 & 0.98 / 5.80 & 0.40 /
  0.49 & \pairrawrate{473/5105}{9.3}{1576/5105}{30.9} & \pairrawrate{40/5105}{0.8}{1166/5105}
  {22.8} & \pairrawrate{855/5105}{16.7}{1311/5105}{25.7} \\
  Whisper-large-v3 & 47.6 / 33.8 & \best{20.7} / 39.9 & 1636 / 4099 & 0.2 / 0.5 & \worst{2.93} / 8.05 &
  0.52 / \worst{1.52} & \pairrawrate{2726/5105}{\worst{53.4}}{2977/5105}{58.3} &
  \pairrawrate{258/5105}{\worst{5.1}}{2285/5105}{44.8} & \pairrawrate{427/5105}{8.4}{442/5105}
  {8.7} \\
  Whisper-large-v3-turbo & 49.4 / \worst{32.5} & 22.8 / 33.7 & 1518 / 4522 & 0.9 / 4.2 & 2.62 /
  \worst{8.41} & 0.48 / 1.35 & \pairrawrate{2389/5105}{46.8}{2989/5105}{\worst{58.6}} &
  \pairrawrate{126/5105}{2.5}{2306/5105}{\worst{45.2}} & \pairrawrate{448/5105}{8.8}{291/5105}
  {5.7} \\
  Distil-Whisper-large-v3 & 48.5 / 39.1 & \worst{107.3} / 140.8 & \worst{5017} / 6911 & 9.1 / 8.2 &
  1.18 / 4.22 & 1.10 / 1.29 & \pairrawrate{857/5105}{16.8}{1570/5105}{30.8} & \pairrawrate{156/5105}
  {3.1}{1211/5105}{23.7} & \pairrawrate{2604/5105}{\worst{51.0}}{2074/5105}{\worst{41.3}} \\
  Qwen2-Audio-7B-Instruct & 49.0 / 66.9 & 58.3 / 122.2 & 2159 / 3743 & \worst{23.6} / \worst{28.9} &
  1.10 / 0.76 & 1.17 / 0.98 & \pairrawrate{128/5105}{2.5}{5/5105}{0.1} & \pairrawrate{0/5105}
  {\best{0.0}}{2/5105}{\best{0.0}} & \pairrawrate{710/5105}{13.9}{1026/5105}{20.1} \\
  Qwen2.5-Omni-3B & \worst{45.8} / 57.3 & 50.7 / 48.8 & 2181 / 2773 & 23.4 / 15.3 &
  1.43 / 1.34 & \worst{1.36} / 1.42 & \pairrawrate{970/5105}{19.0}{174/5105}{3.4} &
  \pairrawrate{5/5105}{0.1}{51/5105}{1.0} & \pairrawrate{1016/5105}{19.9}{378/5105}{7.4} \\
  Qwen2.5-Omni-7B & 59.8 / 64.6 & 39.8 / 46.1 & 1393 / 2784 & 5.4 / 5.0 & 0.99 / 1.63 &
  0.89 / 1.23 & \pairrawrate{419/5105}{8.2}{291/5105}{5.7} & \pairrawrate{5/5105}{0.1}{128/5105}
  {2.5} & \pairrawrate{1419/5105}{27.8}{607/5105}{11.9} \\
  DeSTA2.5-Audio-LLaMA-8B & 46.4 / 66.2 & 41.6 / 32.6 & 4232 / \worst{8992} & 2.0 / 2.7 &
  1.31 / 1.05 & 1.23 / 1.14 & \pairrawrate{20/5105}{\best{0.4}}{5/5105}{\best{0.1}} &
  \pairrawrate{4/5105}{0.1}{0/5105}{0.0} & \pairrawrate{352/5105}{6.9}{398/5105}{7.8} \\
  \hline
  \end{tabular}
  }
  \end{table*}

\begin{table*}[!t]
  \centering
  \scriptsize
  \caption{Whisper-tiny timestamp correction on the self-built gap test sets. Each paired metric reports gap / long-gap. Halluc. rate is file-level.}
  \label{tab:target-main}
  \resizebox{\textwidth}{!}{
  \begin{tabular}{lccccccccc}
  \hline
  Method & mIoU \% $\uparrow$ & MER \% $\downarrow$ & AAS (ms) $\downarrow$ & MAL \% $\downarrow$ & Start MAE (s) $\downarrow$ & End MAE (s) $\downarrow$ & Drift$>$5s \% $\downarrow$ & Drift$>$10s \% $\downarrow$ & Halluc. \% $\downarrow$ \\
  \hline
  Base Whisper-tiny & 63.0 / \worst{38.7} & 79.9 / \worst{347.2} & 951 / \worst{4806} & \worst{5.4} / \worst{17.5} & 1.30 / \worst{7.34} & 0.46 / 0.28 & \pairrawrate{777/5105}{\worst{15.2}}{1883/5105}{36.9} & \pairrawrate{19/5105}{\worst{0.4}}{1365/5105}{26.7} & \pairrawrate{664/5105}{13.0}{982/5105}{19.2} \\

  SFT & 92.7 / 93.5 & 38.3 / 52.2 & 72 / 186 & 0.2 / 0.3 & 0.15 / 0.27 & 0.21 / 0.26 & \pairrawrate{25/5105}{0.5}{102/5105}{2.0} & \pairrawrate{0/5105}{\best{0.0}}{74/5105}{1.4} & \pairrawrate{323/5105}{6.3}{103/5105}{2.0} \\
  SFT timestamp-only & \worst{53.0} / 70.0 & \worst{100.8} / 144.9 & 289 / 273 & 0.3 / 0.4 & 1.45 / 0.16 & \worst{2.32} / 0.81 & \pairrawrate{255/5105}{5.0}{45/5105}{0.9} & \pairrawrate{1/5105}{0.0}{35/5105}{0.7} & \pairrawrate{1997/5105}{\worst{39.1}}{1620/5105}{\worst{31.7}} \\
  SFT decoder & \best{95.3} / \best{96.3} & \best{35.7} / \best{43.4} & \best{46} / \best{37} & \best{0.1} / \best{0.1} & \best{0.07} / \best{0.03} & \best{0.11} / \best{0.06} & \pairrawrate{9/5105}{\best{0.2}}{0/5105}{\best{0.0}} & \pairrawrate{0/5105}{\best{0.0}}{0/5105}{\best{0.0}} & \pairrawrate{340/5105}{6.7}{25/5105}{\best{0.5}} \\
  Edited-time replay & 58.2 / 47.4 & 73.2 / 306.0 & 952 / 3840 & 1.3 / 4.9 & \worst{1.63} / 5.93 & 0.83 / \worst{2.74} & \pairrawrate{446/5105}{8.7}{2208/5105}{43.3} & \pairrawrate{12/5105}{0.2}{1651/5105}{32.3} & \pairrawrate{654/5105}{12.8}{974/5105}{19.1} \\
  Edited-time replay decoder & 63.7 / 55.5 & 74.0 / 304.8 & 743 / 3081 & 1.3 / 4.7 & 1.36 / 4.84 & 0.72 / 2.44 & \pairrawrate{339/5105}{6.6}{1817/5105}{35.6} & \pairrawrate{16/5105}{0.3}{1374/5105}{26.9} & \pairrawrate{693/5105}{13.6}{920/5105}{18.0} \\
  Reduced Teacher Forcing & 93.0 / 94.8 & 39.0 / 45.9 & 84 / 96 & 0.3 / \best{0.1} & 0.15 / 0.13 & 0.24 / 0.11 & \pairrawrate{30/5105}{0.6}{41/5105}{0.8} & \pairrawrate{4/5105}{0.1}{32/5105}{0.6} & \pairrawrate{309/5105}{\best{6.1}}{41/5105}{0.8} \\
  Replayed Reduced Teacher Forcing & 58.2 / 47.2 & 76.3 / 304.0 & \worst{961} / 3869 & 1.4 / 5.0 & \worst{1.63} / 5.99 & 0.81 / 2.73 & \pairrawrate{439/5105}{8.6}{2226/5105}{\worst{43.6}} & \pairrawrate{9/5105}{0.2}{1667/5105}{\worst{32.7}} & \pairrawrate{655/5105}{12.8}{971/5105}{19.0} \\
  \textbf{REDDIT Stage 1 (Ours)} & 93.0 / 94.6 & 45.1 / 56.6 & 69 / 74 & 0.4 / 0.4 & 0.16 / 0.07 & 0.22 / 0.13 & \pairrawrate{22/5105}{0.4}{18/5105}{0.4} & \pairrawrate{3/5105}{0.1}{14/5105}{0.3} & \pairrawrate{423/5105}{8.3}{102/5105}{2.0} \\
  \textbf{REDDIT Full (Ours)} & 93.3 / 95.0 & 45.2 / 53.9 & 66 / 66 & 0.3 / 0.3 & 0.15 / 0.06 & 0.22 / 0.11 & \pairrawrate{22/5105}{0.4}{11/5105}{0.2} & \pairrawrate{4/5105}{0.1}{9/5105}{0.2} & \pairrawrate{422/5105}{8.3}{91/5105}{1.8} \\
  \hline
  \end{tabular}
  }
  \end{table*}

  \begin{table*}[!t]
  \centering
  \scriptsize
  \caption{Out-of-domain timestamp correction and ASR retention. Each paired entry reports mixed-gap / no-gap. Mixed-gap combines single-utterance long-gap and multi-utterance
  gap samples. AAS is sample-weighted and reported only for the mixed-gap evaluation.}
  \label{tab:ood-gap-forgetting_all}
  \setlength{\tabcolsep}{3pt}
  \resizebox{\textwidth}{!}{
  \begin{tabular}{lccccc}
  \hline
  Method
  & AAS (ms) $\downarrow$
  & CV-en MER \% $\downarrow$
  & ASCEND-zh MER \% $\downarrow$
  & ASCEND-en MER \% $\downarrow$
  & ASCEND-mixed MER \% $\downarrow$ \\
  \hline
  Base Whisper-tiny
  & \worst{2752} / --
  & 49.1 / 37.0
  & \worst{136.9} / \best{53.9}
  & 140.6 / 45.5
  & 126.7 / 69.2 \\

  SFT
  & 280 / --
  & 89.1 / 61.6
  & 61.8 / 58.5
  & 299.2 / 303.8
  & 85.7 / 79.9 \\

  SFT timestamp-only
  & 311 / --
  & 55.0 / \best{34.7}
  & 106.1 / 76.5
  & 121.7 / \best{44.4}
  & 110.0 / 90.0 \\

  SFT decoder
  & 315 / --
  & \worst{362.4} / \worst{524.2}
  & 49.9 / 70.5
  & 324.5 / \worst{710.5}
  & 92.6 / \worst{120.3} \\

  Edited-time replay
  & 2733 / --
  & 48.6 / 37.5
  & 130.9 / 56.9
  & 132.1 / 45.8
  & 139.6 / 73.6 \\

  Edited-time replay decoder
  & 2599 / --
  & 48.3 / 36.1
  & 121.8 / 56.0
  & 115.3 / 45.9
  & 132.2 / 71.8 \\

  Reduced Teacher Forcing
  & 301 / --
  & 241.4 / 105.6
  & 69.1 / \worst{166.9}
  & \worst{422.2} / 450.5
  & 97.9 / 112.4 \\

  Replayed Reduced Teacher Forcing
  & 2734 / --
  & 49.6 / 39.8
  & 132.1 / 54.0
  & 135.0 / 45.2
  & \worst{144.3} / 64.5 \\

  \textbf{REDDIT Stage 1 (Ours)}
  & 228 / --
  & \best{38.7} / 38.9
  & \best{48.1} / 58.9
  & 59.7 / 47.2
  & \best{65.7} / \best{58.0} \\

  \textbf{REDDIT Full (Ours)}
  & \best{223} / --
  & 38.8 / 41.3
  & 58.2 / 63.5
  & \best{59.5} / 47.2
  & 70.2 / 60.1 \\
  \hline
  \end{tabular}
  }
  \end{table*}

  \begin{table*}[!t]
  \centering
  \small
  \caption{REDDIT model-scale ablation. The vertical line separates in-domain metrics (left) from OOD metrics (right). Target (in-domain) metrics are gap /
  long-gap; OOD metrics are mixed-gap / no-gap. CV: Common-Voice; AS: ASCEND. CV and AS columns report MER \%.}
  \label{tab:reddit-scale-ablation}
  \setlength{\tabcolsep}{3pt}
  \resizebox{\textwidth}{!}{
  \begin{tabular}{lcccc|cccc}
  \hline
  Method & mIoU \% $\uparrow$ & MER \% $\downarrow$ & AAS (ms) $\downarrow$ & Drift$>$10s \% $\downarrow$ & CV-en \% $\downarrow$ & AS-zh \% $\downarrow$ &
  AS-mix \% $\downarrow$ & OOD AAS $\downarrow$ \\
  \hline
  Whisper-tiny
  & 63.0 / 38.7
  & 79.9 / 347.2
  & 951 / 4806
  & 0.4 / 26.7
  & 49.1 / \best{37.0}
  & 136.9 / \best{53.9}
  & 126.7 / 69.2
  & 2752 / 199 \\

  \textbf{REDDIT Stage 1}
  & 93.0 / 94.6
  & \best{45.1} / 56.6
  & 69 / 74
  & \best{0.1} / 0.3
  & \best{38.7} / 38.9
  & \best{48.1} / 58.9
  & \best{65.7} / \best{58.0}
  & 228 / 144 \\

  \textbf{REDDIT Full}
  & \best{93.3} / \best{95.0}
  & 45.2 / \best{53.9}
  & \best{66} / \best{66}
  & \best{0.1} / \best{0.2}
  & 38.8 / 41.3
  & 58.2 / 63.5
  & 70.2 / 60.1
  & \best{223} / \best{142} \\
  \hline
  Whisper-large-v3
  & 47.6 / 33.8
  & \best{20.7} / 39.9
  & 1636 / 4099
  & 5.1 / 44.8
  & 17.0 / 17.6
  & \best{24.9} / 18.9
  & \best{23.1} / \best{19.9}
  & 2393 / \best{121} \\

  \textbf{REDDIT large-v3 Stage 1}
  & 81.4 / 83.3
  & 37.2 / 49.0
  & 658 / 462
  & 3.1 / \best{0.4}
  & 16.9 / 15.9
  & 36.7 / 19.9
  & 36.0 / 21.3
  & \best{560} / 141 \\

  \textbf{REDDIT large-v3 Full}
  & \best{81.9} / \best{84.0}
  & 21.9 / \best{28.5}
  & \best{329} / \best{337}
  & \best{0.7} / \best{0.4}
  & \best{15.0} / \best{15.7}
  & 25.6 / \best{18.7}
  & 25.2 / 20.9
  & 567 / 123 \\
  \hline
  \end{tabular}
  }
  \end{table*}
\subsection{Non-Speech Gaps Expose Timestamp Drift}

Table~\ref{tab:diagnostic-main} shows that transcript quality does not guarantee temporal grounding. Whisper-family models often produce recognizable text while long-gap start MAE and large-drift rates increase sharply; for example, Whisper-tiny mIoU drops from 63.0\% to 38.7\%, while start MAE rises from 1.30~s to 7.34~s. The strongest lexical systems are not automatically the strongest temporal systems: Whisper-large-v3 has the best gap MER among Whisper models, but it also has the worst gap Drift$>$5s rate. These results support evaluating timestamp drift as an acoustic-grounding error rather than as a byproduct of transcript quality. The table also separates drift from hallucination: large temporal displacements measure misplaced supported speech, whereas hallucination rate measures unsupported loops or boilerplate.

\subsection{REDDIT Corrects Timestamp Drift}

Table~\ref{tab:target-main} evaluates Whisper-tiny interventions. SFT decoder tuning obtains the best target timestamp accuracy, but it does so by updating a much larger decoder scope and later degrades severely on OOD ASR. The restricted-scope baselines show that parameter efficiency alone is not enough: SFT timestamp-only, edited-time replay, and replayed RTF leave substantial long-gap drift. REDDIT Full reaches SFT-level temporal correction while keeping the replay-conditioned Stage-1 edit as its first step: long-gap mIoU improves from 38.7\% to 95.0\%, AAS from 4806~ms to 66~ms, and Drift$>$10s from 26.7\% to 0.2\%. It also reduces long-gap hallucination rate from 19.2\% to 1.8\%, which we treat as an auxiliary non-speech robustness effect rather than the definition of drift. The small gap between REDDIT Stage 1 and REDDIT Full indicates that replay-conditioned editing already supplies the main correction signal, while Stage 2 mainly consolidates the corrected prefix behavior.

\subsection{OOD Robustness and Retention}

Table~\ref{tab:ood-gap-forgetting_all} jointly reports timestamp correction on mixed-gap OOD audio and recognition retention under both mixed-gap and no-gap conditions. REDDIT
  Full reduces the sample-weighted mixed-gap AAS from 2752 to 223~ms. On the no-gap CV-en and ASCEND-en sets, its MERs remain close to those of the base model (41.3\% versus
  37.0\%, and 47.2\% versus 45.5\%, respectively), whereas decoder SFT reaches 524.2\% and 710.5\%. Under mixed-gap conditions, REDDIT Full also reduces MER relative to the
  base model on all four OOD datasets, while avoiding the severe CV-en and ASCEND-en degradation observed for decoder SFT and RTF. These results support the distribution-
  editing view of timestamp adaptation: timestamp positions should be corrected while non-timestamp behavior remains anchored to the base model.

 \subsection{Transfer to Natural Broadcast Audio}
  \label{sec:natural-transfer}
  Table~\ref{tab:natural-broadcast} evaluates REDDIT on natural broadcast audio containing naturally occurring non-speech intervals. On \emph{The Daily Show}, REDDIT reduces
  AAS from 3010 to 1252~ms and MER from 30.8\% to 24.4\%. SFT decoder achieves a comparable AAS of 1259~ms but increases MER to 92.2\%, showing that timestamp correction alone
  does not ensure recognition retention. On \emph{The Night Night Show} in Mandarin, REDDIT reduces AAS from 3870 to 2257~ms, the Drift$>$5s rate from 40.9\%
  to 14.9\%, and MER from 71.5\% to 58.2\%. Although SFT decoder achieves a lower AAS of 2008~ms on this dataset, REDDIT yields lower drift and MER. Overall, these results show
  that correction learned from gap-augmented training audio transfers to unspliced broadcasts in both languages while better preserving recognition quality.

\subsection{Ablation Summary}
Table~\ref{tab:reddit-scale-ablation} compares REDDIT Stage 1 and REDDIT Full across Whisper-tiny and Whisper-large-v3.
The target and OOD tables isolate decoder context, replay, and reduced teacher forcing. Teacher-forced RTF is strong on the target timestamp set, but degrades on mixed-gap OOD recognition. Replayed RTF shows that adding replay to an In-Sync-inspired objective is insufficient by itself: without the non-timestamp KL anchor, the model can still move away from its base lexical behavior. REDDIT's Stage-1 advantage is the combination of cached replay context, edited timestamp targets, and KL retention on non-timestamp positions.

Stage 1 isolates the core replay-conditioned timestamp editing effect, while REDDIT Full adds the edited-prefix refinement used by the complete pipeline. On Whisper-tiny, Stage 2 gives modest additional timing and OOD alignment gains; on Whisper-large-v3, the REDDIT rows show that the two-stage schedule is more important for scaling. The Stage-1 large-v3 row already repairs much of the temporal displacement, but REDDIT Full improves target long-gap MER and most OOD MER values after the Stage-1 checkpoint, while mixed-gap OOD AAS remains close to Stage 1 and far below the base model. This suggests that edited-prefix refinement helps larger decoders use the corrected timestamp trajectory while retaining the Stage-1 replay anchor.

% \FloatBarrier
\section{Conclusion}
\label{sec:conclusion}

We studied non-speech-induced drift in model-generated timestamps: autoregressive ASR systems can transcribe recognizable speech while placing it on the wrong part of the audio timeline. Controlled gap and long-gap evaluations across timestamp-producing ASR and audio-language systems show that this error is distinct from transcript error and hallucination. Naive timestamp-corrected fine-tuning can repair target alignment but causes severe forgetting on non-target ASR data, motivating timestamp adaptation as a distribution-editing problem. REDDIT addresses this with a two-stage pipeline: replay-conditioned timestamp editing preserves the frozen base distribution on non-timestamp tokens, and a short edited-prefix refinement consolidates the corrected timestamp behavior. On Whisper-tiny, this pipeline improves long-gap timestamp robustness and mixed-gap OOD alignment without adding inference-time VAD, forced alignment, or post-processing. On Whisper-large-v3, the Stage-2 refinement becomes more important, suggesting that REDDIT is a scalable strategy for preserving recognition while editing generated timestamps.

\subsection{Future Work}

Our intervention study focuses on explicit timestamp-token ASR, where token-level editing is well defined. Extending REDDIT to audio-language models is more complex because temporal information may be expressed in free-form text and non-ASR abilities such as speech instruction following and audio reasoning must be preserved. Future work should evaluate forgetting-resistant temporal editing for LALMs under broader protocols, including Speech-IFEval-style tests \cite{speech_ifeval}, and analyze how timestamp editing changes cross-attention and acoustic grounding.
\newpage
\FloatBarrier
% \input{sections/07_acknowledgment}
% \clearpage

\end{document}